# Extraction in Dutch with Lexical Rules

Gerrit Rentier*
Institute for Language Technology and Artificial Intelligence
Tilburg University, PO Box 90153, 5000 LE Tilburg, The Netherlands
rentier@kub.nl

**Abstract**

Dutch argument extraction is analyzed in the version of Head-driven Phrase Structure Grammar which is sketched in [Pollard&Sag(1994)], chapter 9. In this version of HPSG the existence of *traces* is denied and instead extraction information is introduced by *lexical rules* operating on the features of lexical heads. We present such lexical rules to account for Dutch *verb second* and the highly idiosyncratic properties of Dutch *preposition stranding*, thus illustrating the descriptive power of such lexical rules. We also generalize the analysis to account for the behaviour of Dutch P$^o$ w.r.t. neuter pronouns and define P$^o$ in the hierarchical lexicon so that P$^o$ is disallowed to locally govern any pronoun which denotes a neuter referent.

## 1 Dutch Verb Second and Preposition Stranding

Dutch displays an intriguing idiosyncracy with respect to preposition stranding, as can be read from the contrast between (1b) and (1c);

(1) a. Aan welke stichting schenkt Beatrix het huis ?
To which foundation donates Beatrix the house ?
"To which foundation does Beatrix donate the house ?"

b. * Wat schenkt Beatrix het huis aan ?
What[R−] donates Beatrix the house to ?
"What does Beatrix donate the house to ?"

c. Waar schenkt Beatrix het huis aan ?
What[R+] donates Beatrix the house to ?

*This research report is a slightly corrected version of [Rentier(1994b)]. The relevant corrections concern semantic technicalities in Sect. 5 and an erroneous definition in an analysis in Sect. 6. Author was, over the relevant periods of time, sponsored by EC projects ESPRIT P5254 (PLUS) and P6665 (DANDELION), a travel grant from the Netherlands Organization for Scientific Research (NWO) and a travel grant from SHELL NEDERLAND. Thanks for many useful suggestions and for detailed comments on earlier versions of this paper to Bart Geurts, Josée Heemskerk, Paola Monachesi, John Nerbonne, Carl Pollard, Ivan Sag, Chris Sijtsma, Wietske Sijtsma, Craig Thiersch, Margriet Verlinden, attendants of CLIN III, Tilburg, the Netherlands, October '92 for comments on [Rentier(1993)], and attendants of the *First International Workshop on HPSG* in Columbus, Ohio, USA, August '93 (where an earlier version of this paper was read) and the audience of KONVENS94 in Vienna, Austria, September '94.

Fronting of a PP non-subject argument is always possible with the PP as a whole (pied piping, (1a)), modulo the usual island constraints. This is just an instance of verb second, which we shall discuss in Sect. 3. The interesting idiosyncracy however is that preposition stranding is impossible with a certain variant of the WH-pronoun, but grammatical with another ((1b) vs. (1c)). Starting with [van Riemsdijk(1978)] this contrast and similar contrasts with neuter demonstrative, relative and clitic pronouns in Dutch have been explained by the syntactically relevant absence vs. presence of the phoneme /R/.

In sections 4 and 5 we will give a lexicalist account of the above and related contrasts. The account will be lexical in the sense that it will introduce no additional mechanisms, principles or phrase structure rules into HPSG but instead carefully defines the local and nonlocal selection properties of prepositions while introducing minor constraints on three independently motivated lexical rules. In Sect. 6, however, we tentatively suggest an additional immediate dominance schema to account for filler-like constituents in the Mittelfeld (cf. [Rentier(1993)]).

## 2 Subjects, Complements and Dutch Clause Structure

Instead of the feature SUBCAT, which is put forward to list a head's locally selected arguments in [Pollard&Sag(1987)], we will adopt the division of arguments as subjects and nonsubjects which is motivated for English in [Borsley(1987)]. In Chap. 9 of [Pollard&Sag(1994)] this approach to local selection of arguments is developed further and leads to the postulation of the Valence Principle. The Valence Principle refers to the *valence* features SUBJ and COMPS through 'F' in the following definition:

(2) **Valence Principle** [Pollard&Sag(1994)], Chap. 9, pp.348
In a headed phrase, for each valence feature F, the F value of
the head-daughter is the concatenation of the phrase's F value with
the list of SYNSEM values of the F-daughters value.

Of course we want a theory of valence to be universal. If we assume it for English for the reasons given by [Borsley(1987)] and Chap. 9 of [Pollard&Sag(1994)], then we should be able to succesfully implement it in our analysis of Dutch as well. Furthermore, with Dutch this division allows for an interesting analysis of Cross Serial Dependencies, as discussed in [Rentier(1994a)].[1] Therefore we assume lexical entries for Dutch finite verbs like *schenkt* ("donates") to look like (3), where we also include the NONLOCAL features that pass on extraction information; the use of these features will be extensively illustrated in the remainder of the paper.

The effect of the Valence Principle on a headed phrasal sign that is headed by, e.g., the lexical sign in (3) is that this sign can only be regarded as "complete" or *saturated* if the lexical sign is combined with the appropriate arguments;

[1] Specifically, the fact that we list the least oblique argument of a governed verb in a separate feature "SUBJ", allows for a particularly intuitive analysis of the fact that, in Dutch cross serial constructions, the governing verb assigns case to that argument of the governed verb, but not to any of the other arguments of the governed verb. For discussion, cf. [Rentier(1994a)].

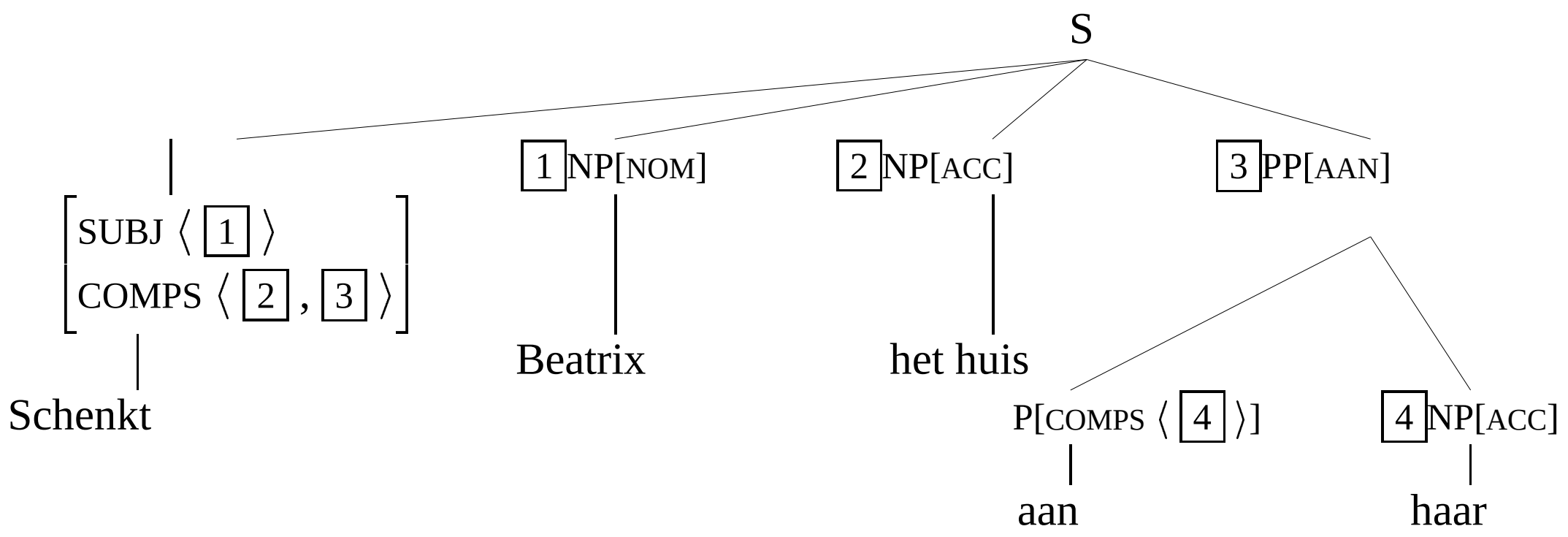

(3)

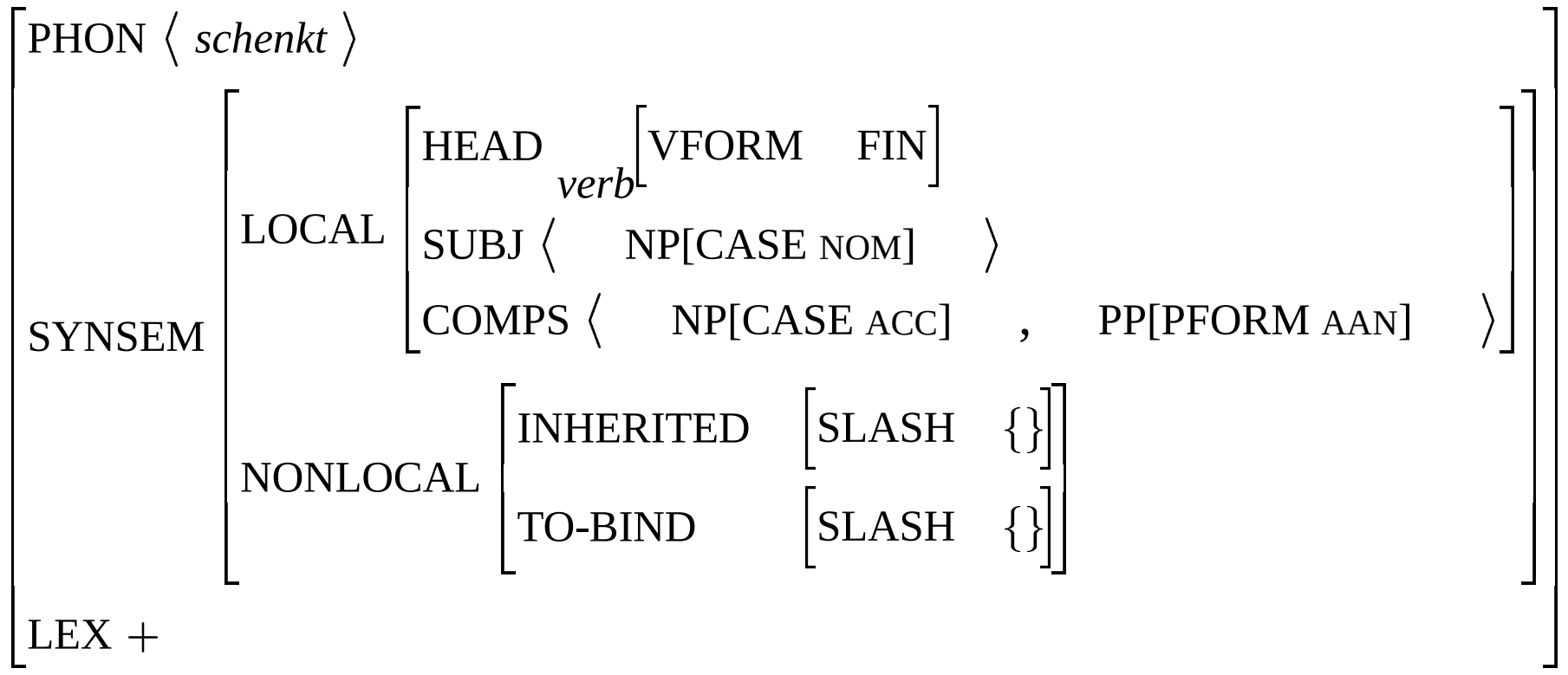

In the case of (3) the ‘appropriate’ arguments are a nominative subject, an accusative object and a PP-complement which is headed by *aan* (“to”). If we assume a flat clause structure analysis ([Pollard(forthc.)], [Nerbonne(1994)]) for Dutch, and assume lexical signs like (3), then the following immediate dominance statements suffice to describe the fragment we are concerned with;[2]

(4) a. (Schema I) a [SUBJ ⟨ ⟩,COMPS ⟨ ⟩] phrase with daughters of sort *head-subj-comps-struc* in which the head-daughter is a lexical sign

b. (Schema II) a [COMPS ⟨ ⟩] phrase with daughters of sort *head-comps-struc* in which the head-daughter is a lexical sign

c. (Schema III) a phrase of sort *filler-head-struc* with a filler-daughter that has a LOCAL value token-identical to both the INHER|SLASH and the TO-BIND|SLASH value of the head-daughter, where the head-daughter is a finite sentence

Together immediate dominance schemata I and II, the demands made by the Valence Principle and the selectional requirements made by the lexical entry for *schenkt* give rise to phrase structure analyses of Dutch yes/no-interrogatives and PP’s as in the figure above.[3]

---

[2] Here *head-subj-comps-struc* indicates that the daughters of the phrase include a head, a subject and complements, not necessarily in that order; *head-comps-struc* indicates the same, but without the subject daughter. Cf. Chap. 9 of [Pollard&Sag(1994)] for detailed discussion.

[3] In this figure and througout the paper, recurring i’s indicate structure sharing, that is token-identity of information, as is common usage in HPSG.

## 3 Verb Second as Argument Extraction without Traces

In Chap. 9 of [Pollard&Sag(1994)] a theory of extraction which does not employ any notion of traces or empty categories is suggested. This theory is further motivated by [Sag&Fodor(1994)], who argue convincingly that no theory external evidence for traces exists. Also, they point out several advantages of such a lexically based theory of extraction, for instance w.r.t. isolated idiosyncracies that invite an analysis as lexical exceptions.

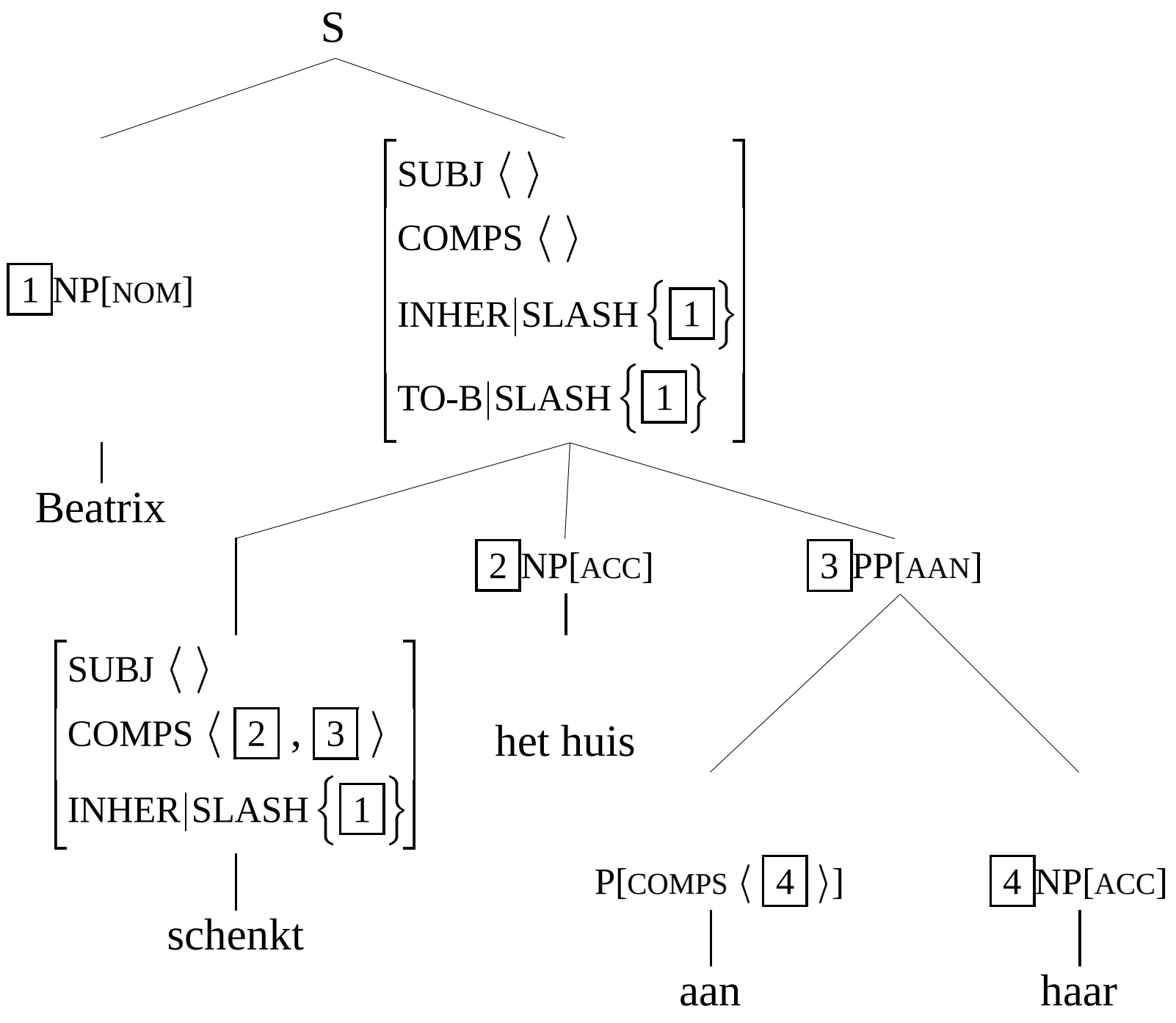

We argue that such a theory, together with the assumptions from Sect. 2, allows for an elegant traceless version of the analysis of verb second in Germanic which is discussed in [Pollard(forthc.)]. Following Pollard, we assume that in HPSG, the verb second phenomenon should be modeled as extraction from a head-initial (flat) clause. In the trace-less model of extraction, we can then describe a sentence with the subject in the Vorfeld by assuming that the Subject Extraction Lexical Rule for Dutch should read as follows;

(5) **Subject Extraction Lexical Rule** *Dutch*

$$\begin{bmatrix} \text{LOC|SUBJ } \langle \boxed{1} \rangle \\ \text{NONLOC|INHER|SLASH } \{\ \} \end{bmatrix} \Rightarrow \begin{bmatrix} \text{LOC|SUBJ } \langle \rangle \\ \text{NONLOC|INHER|SLASH } \{ \boxed{1} \} \end{bmatrix}$$

A declarative Dutch sentence with an unmarked theme will then have a structure as in the figure above, where the ‘top’ is licensed by ID-Schema III (cf. (4c)) and the ‘middle’ is licensed by ID-Schema II (cf. (4b)). The SELR, listed in (5), has applied to the entry in (3) to produce the homophonous entry for *schenkt* which is the head in the tree above.

Any value of the feature ‘INHER’ (for “INHERITED”) is subject to the Nonlocal Feature Principle:

(6) **Nonlocal Feature Principle** [Pollard&Sag(1994)], Chap. 4, pp.164
For each NONLOCAL feature, the INHER value on the mother is the union of the INHER values on the daughters minus the TO-BIND value on the head-daughter

The effect of the NFP is that the nonlocal selection information percolates 'up' in the structure until it can be associated with an appropriate 'filler'. This appropriateness is forced by the demand of token-identity mentioned in the immediate dominance schema which combines the filler with the clause which it is extracted from, that is, Schema III.

This approach also allows for a traceless analysis of non-subject arguments in the Vorfeld; we will discuss this in its relation to Dutch preposition stranding.

# 4 A Traceless Account of Dutch Preposition Stranding

As discussed in the introduction, Dutch displays an interesting idiosyncracy with respect to extraction from prepositional phrases, illustrated in (1) with the contrast between (1b) and (1c).[4] Such contrasts are usually attributed to the presence vs. the absence of the phoneme /R/ in a syntactically relevant way. If we indicate this property as [R+] and [R−] on the appropriate lexical items, we might account for this idiosyncracy by adopting a modified version of the Complement Extraction Lexical Rule (cf. Chap. 9, [Pollard&Sag(1994)]) for Dutch prepositional heads;

(7) **Preposition Complement Extraction Lexical Rule** *First Version*

$$\begin{bmatrix} \text{LOC}|\text{HEAD} \ \ \textit{prep} \\ \text{LOC}|\text{COMPS} \ \langle \boxed{1}[\text{R}-] \rangle \\ \text{NONLOC}|\text{INHER}|\text{SLASH} \ \{ \ \} \end{bmatrix} \Rightarrow \begin{bmatrix} \text{LOC}|\text{HEAD} \ \ \textit{prep} \\ \text{LOC}|\text{COMPS} \ \langle \rangle \\ \text{NONLOC}|\text{INHER}|\text{SLASH} \ \{ \boxed{1}[\text{R}+] \} \end{bmatrix}$$

This lexical rule is restricted by the occurence of [R+] in such a way that the introduced extraction information on a PP will always concern an extracted element which is [R+].[5] Through (7) we can account for the contrast between (1c) and (1b) in a straightforward manner; we merely have to state in the lexicon that *waar* is [R+] and that *wat* is [R−]. The analysis of (1c) is given in the figure next page (where $P^o$ is derived through the PCELR).

Note furthermore that with (7) we propose a substantial extension to the traceless theory of extraction through lexical rules. This is so since here we place idiosyncratic restrictions not on the head which licenses the unbounded dependency, but on the element which is extracted itself.[6]

To allow for (8a), and extraction of non-subject arguments of verbs in general, Dutch must have a separate Complement Extraction Lexical Rule for verbs. As we can see from the contrast with (8b), verbs can only be nonlocally related to pronouns which are [R−]:[7]

---

[4] For an extensive discussion of data, cf. a.o. [van Riemsdijk(1978)].

[5] We assume that in the lexicon, all NP arguments which are on COMPS are [R−] accounting for "*Beatrix waardeert dat/*daar*". Further restrictions on NP arguments come into play with $P^o$, cf. Sect. 5.

[6] That is, we impose other restrictions than the syntactic and semantic selection restrictions which are standardly imposed by the lexical head which selects the argument.

[7] N.B.; The underscores in (8) and throughout this paper are *not* to be understood as traces or any other kind of empty categories. The underscores indicate, for expository reasons, positions where an argument could have been locally realized, but isn't locally realized because it is instead nonlocally realized elsewhere.

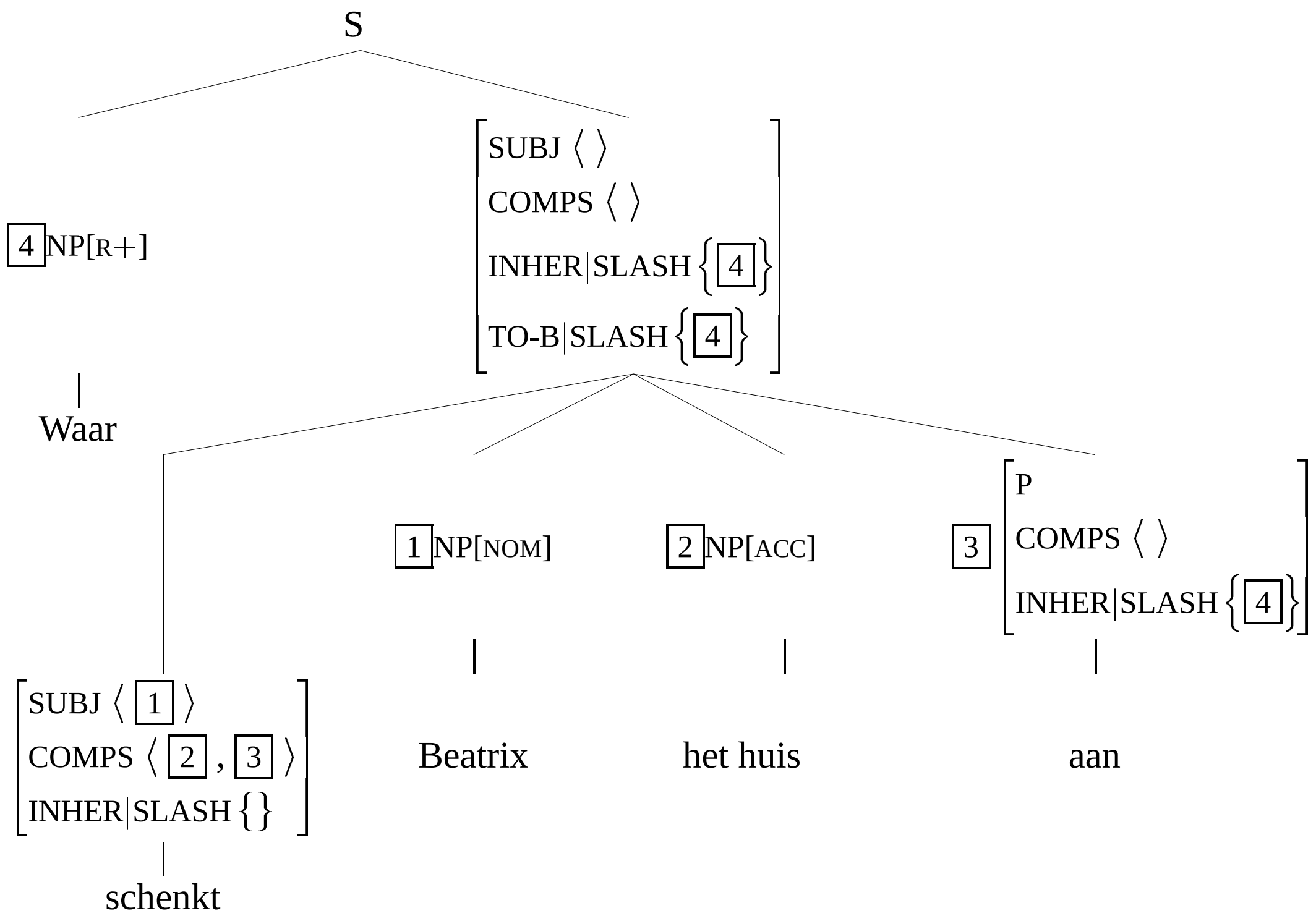

(8) a. Dat₁ zal Peggy waarschijnlijk __₁ waarderen
That[R−] will Peggy probably appreciate
"Peggy will probably appreciate that"

b. * Daar₁ zal Peggy waarschijnlijk __₁ waarderen
That[R+] will Peggy probably appreciate

c. Daar₁ zal Peggy waarschijnlijk [op __₁] rekenen
That[R+] will Peggy probably on count
"Peggy will probably count on that"

Topicalization of an [R+] demonstrative pronoun can obviously only be licensed by P°, cf. (8c). Therefore, we propose that in Dutch, any lexical rule which introduces an unbounded dependency on a verb will constrain the introduced element to be [R−]; this is reflected in the Dutch rule for verb complement extraction in (9).

(9) **Verb Complement Extraction Lexical Rule**

$$\begin{bmatrix} \text{LOC|HEAD } \textit{verb} \\ \text{LOC|COMPS } \langle \ldots, \boxed{1}[\text{R}-], \ldots \rangle \\ \text{NONLOC|INHER|SLASH } \{\ \} \end{bmatrix} \Rightarrow \begin{bmatrix} \text{LOC|HEAD } \textit{verb} \\ \text{LOC|COMPS } \langle \ldots\ldots \rangle \\ \text{NONLOC|INHER|SLASH } \{ \boxed{1}[\text{R}-] \} \end{bmatrix}$$

The same constraint should and can be built into the Dutch SELR (5).

## 5 A Generalization of the Analysis

On our account so far, all examples in (10) should be grammatical;

(10) a. Hij heeft op het slechte weer gerekend
He has on the bad weather counted

b. Hij heeft op hem/ haar gerekend
He has on him[R−]/ her[R−] counted

c. * Hij heeft op het/ daar/ dat gerekend
He has on it[R−]/ that[R+]/ that[R−] counted

However, the facts in (10) seem to generalize to the observation that except for full NPs, only pronouns with male or female gender can be locally governed by $P^o$. In the ungrammatical constructions given in (10c) the pronouns denote referents of neuter gender. In [Pollard&Sag(1994)], it is assumed that non-predicative $P^o$ is semantically vacuous. Consequently, the CONTENT of a projection of $P^o$ is structure shared with the CONTENT-value of its NP-argument.[8] From (10), it seems then that a Dutch preposition with a CONTENT-value of the type of a pronoun which has NEUTER as the value of GENDER cannot govern that pronoun locally. Instead, such a $P^o$ should always introduce an extraction and thus select its argument nonlocally instead of locally.[9]

Here we propose to capture this generalization by, firstly, imposing a negative constraint on the semantics of non-predicative $P^o$ in the hierarchical lexicon for Dutch:[10]

(11)
$$\begin{bmatrix} \text{PHON} & \langle \text{X} \rangle \\ \text{SYNSEM} & \begin{bmatrix} \text{LOCAL} & \begin{bmatrix} \text{CATEGORY} & \begin{bmatrix} \text{HEAD} & {}_{prep}\begin{bmatrix}\text{PFORM} & \text{X'}\end{bmatrix} \\ \text{COMPS} & \langle\ \text{NP[ACC][R}-\text{]:}\boxed{1}\ \rangle \end{bmatrix} \\ \text{CONTENT} & \boxed{1}\ \textbf{NOT}\ {}_{ppro}\begin{bmatrix}\text{INDEX} & \begin{bmatrix}\text{GENDER NEUTER}\end{bmatrix}\end{bmatrix} \end{bmatrix} \\ \text{NONLOCAL} & \begin{bmatrix} \text{INHERITED} & \begin{bmatrix}\text{SLASH}\{\}\end{bmatrix} \\ \text{TO-BIND} & \begin{bmatrix}\text{SLASH}\{\}\end{bmatrix}\end{bmatrix} \end{bmatrix} \end{bmatrix}$$

Without working out the details, we assume that we have a hierarchical lexicon which eliminates redundancies through inheritance, as suggested in Chap. 8 of [Pollard&Sag(1987)]. We organize the lexicon so that all lexical entries for $P^o$ are as in (11), so that they can license any complement locally as a value of COMPS if it is an accusative NP which is [R−] and not a neuter pronoun.

Such entries are appropriate for any Dutch non-predicative $P^o$ which locally governs a complement which is neuter but not pronominal (cf. (10a)) or pronominal but not neuter

[8] The feature CONTENT of any nominal object ranges over *nom-obj* or a subtype of it; one such subtype is *ppro*. Cf. [Pollard&Sag(1994)], Chap. 1, pp.24-26, for discussion of further details.

[9] Such nonlocal selection of $P^o$ is always for a neuter pronoun; cf. the examples in (1c), (8c), (14b) and, on our tentative analysis presented in Sect. 6, also in (16).

[10] In (11), (12) and (13), the colon, “:”, abbreviates the path SYNSEM|LOC|CONTENT.

(cf. 10b)). The negative constraint on the value of the CONTENT feature correctly excludes ungrammatical constructions involving neuter pronoun objects, like (10c).

Furthermore, entries for P$^{o}$ like (11) correspond to a PP which *must* locally select its argument; no P$^{o}$ with semantics like (11) will be allowed to appear in a 'slashed' form. This is guaranteed because, in the lexical rule approach to extraction, categories only appear slashed if a lexical rule produces them as such. Since any PP licensed as an instance of (11) is specified as empty for the feature SLASH, no PP licensed by an instance of (11) will be nonlocally related to its complement.

The question as to how we should license grammatical cases of preposition stranding is of course still answered through the Preposition Complement Extraction Lexical Rule. But we should change it in such a manner that it will erase the negative constraint on the value of CONT, thus giving rise to the desired results;

(12) **Preposition Complement Extraction Lexical Rule** *Final*

$$\begin{bmatrix} \text{HEAD} \;\; \textit{prep} \\ \text{COMPS} \; \langle \; \boxed{1}[\text{R}-]{:}\boxed{2} \; \rangle \\ \text{CONT}\,\boxed{2}\,\textbf{NOT}\textit{ppro}[\text{INDEX}[\text{GEND NEUT}]] \\ \text{INHER}|\text{SLASH} \; \{ \; \} \end{bmatrix} \Rightarrow \begin{bmatrix} \text{HEAD} \;\; \textit{prep} \\ \text{COMPS} \; \langle \; \rangle \\ \text{CONT}\,\boxed{3}\;\textit{ppro}[\text{INDEX}[\text{GEND NEUT}]] \\ \text{INHER}|\text{SLASH} \; \{ \; \boxed{1}[\text{R}+]{:}\boxed{3} \; \} \end{bmatrix}$$

Application of this PCELR to entries like in (11) gives rise to entries like (13):

(13)

$$\begin{bmatrix} \text{PHON} \; \langle \; X \; \rangle \\ \text{SYNSEM} \begin{bmatrix} \text{LOCAL} \begin{bmatrix} \text{CATEGORY} \begin{bmatrix} \text{HEAD} \; {}_{\textit{prep}}\begin{bmatrix} \text{PFORM} \;\; \text{X'} \end{bmatrix} \\ \text{COMPS} \; \langle \;\; \rangle \end{bmatrix} \\ \text{CONTENT} \; \boxed{3} \; {}_{\textit{ppro}}\begin{bmatrix} \text{INDEX} \begin{bmatrix} \text{GENDER NEUTER} \end{bmatrix} \end{bmatrix} \end{bmatrix} \\ \text{NONLOCAL} \begin{bmatrix} \text{INHERITED} \begin{bmatrix} \text{SLASH} \; \{ \; \text{NP}[\text{ACC}][\text{R}+]{:}\boxed{3} \; \} \end{bmatrix} \\ \text{TO-BIND} \begin{bmatrix} \text{SLASH} \; \{\} \end{bmatrix} \end{bmatrix} \end{bmatrix} \end{bmatrix}$$

The entries which are like (13) then are assumed to take part in all grammatical cases of preposition stranding (cf., e.g., the tree for (1c); also, (8c), (14b) and (16)).

## 6 Extensions to the Analysis

Similar lexical rules can explain obviously related contrasts between two variants of the Dutch relative pronouns (cf. (14)); the only difference will be that this PCELR should make reference to the nonlocal feature REL (Chap. 5, [Pollard&Sag(1994)]), and not to SLASH.

| | | | | | | | | | | |
|---|---|---|---|---|---|---|---|---|---|---|
| (14) | a. | * | Het | slechte | weer | wat$_1$ | [Peggy | [op $___1$] | heeft | gerekend] |
| | | | The | bad | weather | what[R−] | Peggy | on | has | counted |
| | | | "The spell of rain that Peggy has counted on" | | | | | | | |
| | b. | | Het | slechte | weer | waar$_1$ | [Peggy | [op $___1$] | heeft | gerekend] |
| | | | The | bad | weather | what[R+] | Peggy | on | has | counted |

Furthermore, the Nonlocal Feature Principle, when examined closely,[11] allows fillers to be sisters to the arguments from which they are extracted.

The relevant fact is, that both the filler and the argument(s) from which it is 'extracted' should be allowed as sisters of the head-daughter, by some additional immediate dominance schema;

(15) (Schema IIIb) (Additional)
a phrase of sort *filler-(subj)-comps-head-struc* with a filler-daughter that has a LOC value token-identical to the TO-BIND|SLASH value of the head-daughter and the INHER|SLASH value of some COMPS-daughter, where the head-daughter is a finite lexical verb and the filler is [R+]

This allows for structures where fillers can be at the same level in the tree as the heads to which they are nonlocally related. If we allow for such phrasal structures, then the version of the PCELR in (12) will also account for grammatical constructions in Dutch where the demonstrative pronoun *"daar"* or the clitic pronoun prepositional object *"er"* appear in the Mittelfeld, not the Vorfeld;

(16) a. Peggy hoeft *daar*$_1$/ * *dat*$_1$ niet [op __$_1$] te rekenen
Peggy has that[R+]/ that[R−] not on to count
"Peggy shouldn't count on that"

b. Beatrix schenkt *er*$_1$/ * *het*$_1$ geen huis [aan __$_1$]
Beatrix donates it[R+]/ it[R−] no house to
"Beatrix doesn't donate a house to it"

This analysis, though stipulative,[12] gives a natural account of the relation between the preposition and its pronoun object. Firstly, it explains the discontinuity between the P$^o$ and its object in (16), cf. next page figure. Secondly, it is consistent with the generalization that P$^o$ cannot *locally* govern any neuter pronoun, cf. (10). We claim this is an appealing advantage of the above analysis of Dutch preposition stranding since it allows for a completely unified account of the distribution of R-pronouns in Dutch.[13]

[11] More closely then in [Rentier(1994b)]; in [Rentier(1994b)], we incorrectly assumed that the NFP should be revised in order to account for data such as the data in (16).

[12] It is, of course, a goal of the theory to reduce and not increase the number of immediate dominance schemata. Perhaps it is technically feasible to construe IIIb in (15) as a phrasal subtype which multiply inherits from the phrasal types I, II and III in (4).

[13] Further discussion of other empirical arguments for such a unification of the account: [Rentier(1993)].

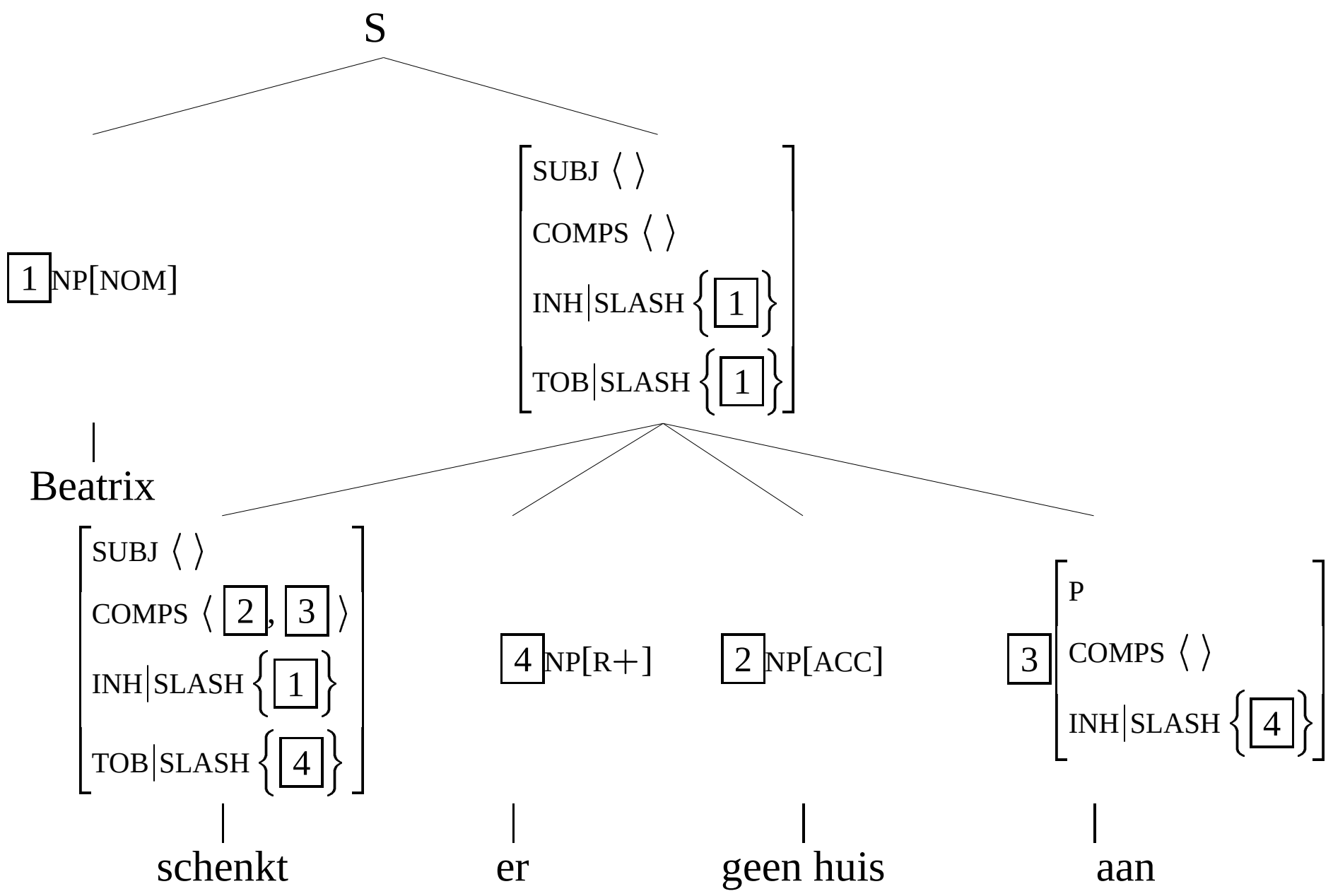